# Fast Segmentation of Left Ventricle in CT Images by Explicit Shape Regression using Random Pixel Difference Features

Peng Sun, Haoyin Zhou, Devon Lundine, James K. Min, and Guanglei Xiong

Dalio ICI, Weill Cornell Medical College, Cornell University, USA
{pes2021,haz2011,del2014,jkm2001,gux2003}@med.cornell.edu

**Abstract.** Recently, machine learning has been successfully applied to model-based left ventricle (LV) segmentation. The general framework involves two stages, which starts with LV localization and is followed by boundary delineation. Both are driven by supervised learning techniques. When compared to previous non-learning-based methods, several advantages have been shown, including full automation and improved accuracy. However, the speed is still slow, in the order of several seconds, for applications involving a large number of cases or case loads requiring real-time performance. In this paper, we propose a fast LV segmentation algorithm by joint localization and boundary delineation via training explicit shape regressor with random pixel difference features. Tested on 3D cardiac computed tomography (CT) image volumes, the average running time of the proposed algorithm is 1.2 milliseconds per case. On a dataset consisting of 139 CT volumes, a 5-fold cross validation shows the segmentation error is $1.21 \pm 0.11$ for LV endocardium and $1.23 \pm 0.11$ millimeters for epicardium. Compared with previous work, the proposed method is more stable (lower standard deviation) without significant compromise to the accuracy.

## 1 Introduction

The Left Ventricle (LV) has the thickest wall of the heart's four chambers for pumping blood throughout the body. LV myocardium consists of endocardial (inner) and epicardial (outer) layers, which can be visualized by cardiac CT/MR imaging. An accurate segmentation is the first step to assess LV shape, function, and perfusion from imaging data. Compared to traditional unsupervised approaches (e.g. region growing, deformable models, active shape models (ASM), atlas based registration), learning-based segmentation in the seminal work [10][4][11] is more robust and efficient, where the segmentation algorithm has two separate steps: heart localization and boundary delineation. Although considerably faster than previous methods, the learning-based segmentation is unable to work in real time due to a large number of expensive searches, including both the global search for heart localization where all voxels of the volume are scanned and the local search for boundary delineation where every point of the surface model is adjusted.

In this paper, however, we combine the two steps and propose a novel joint localization-delineation algorithm, which can segment LV from a cardiac CT volume in millisecond level (average of 1.20 ms in experiments) without noticeable compromise of accuracy. In contrast, the previous state-of-the-art method takes typically several seconds. The speedup is mainly due to the following two aspects of our algorithm:

1. A direct shape regression. The LV is modeled by a set of points, then a regressor is trained, taking the whole CT image as input to predict the positions of all LV points. This is inspired by the observation that each sub volume patch has definite anatomical structure and thus "encodes" the true position of the organ interested [2]. Unlike simply training a regression tree for the positions of specific organs [2], we adopt the Cascaded Pose Regression [3][1] to capture LV's complicated shape variation. In this way, we avoid the time consuming sliding-window search for LV localization [10] that the Marginal Space Learning technique attempts to improve [10], where the correlation among sub patches is ignored due to an independent prediction at every patch. Previous studies also leverage the inter-patch correlation by taking a voting based regression [12] or Generalized Hough Transform based strategy [4] for LV localization, however, an extra boundary delineation step is still needed.

2. A novel underlying feature. In this study, the LV shape regressor is fed with Random Pixel Difference [8] feature that is efficient yet effective. In previous studies [10] [11], Haar-like feature [9] is adopted, but, it is computationally expensive despite using the speedup trick of Integral Image, as it involves many sub image patches' averaging and mutual subtraction. Meanwhile, the image edge based feature utilized by Ecabert et al [4] is also time consuming. Both the aforementioned features are, we argue, "over-qualified" to discriminate the LV boundary whose example is shown in Fig.1. Note the regular pixel values distribution, the Random Pixel Difference feature is hopefully already a strong signature for LV boundary against other image patches. In addition, it is computationally much cheaper than previously mentioned features.

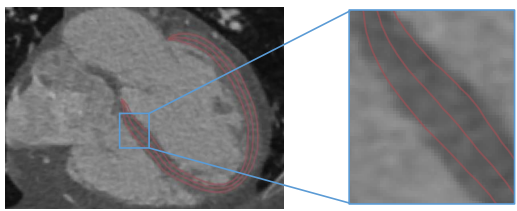

**Fig. 1.** The distribution of pixel values along the boundaries.

In the rest of this manuscript, we will describe the method in Section 2 and discuss the experimental results in Section 3. Finally, we conclude in Section 4 and discuss the future work.

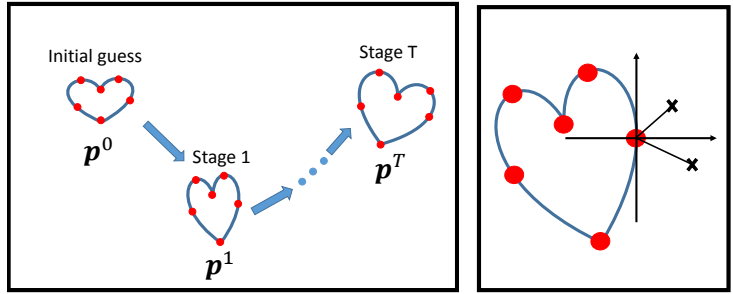

**Fig. 2.** Illustration of the proposed algorithm. In this example the number of landmarks $L = 6$, the number of features per landmark $M = 1$. **Left**: the multi stage shape updating of the Cascaded Pose Estimation; **Right**: The Random Pixel Difference feature for a landmark by subtracting the pixel values of the randomly generated point-pair. Up to $ML = 6$ features will be fed into the stage regressor in this example. See the ptext in Section 2 for explanations.

## 2   Method

A quick overview of the proposed algorithm is in Fig.2. Greater details are given in the following. The LV is modeled by a set of $L \in \mathbb{N}^+$ landmark points, where the first half accounts for the inner surface while the second half for the outer surface. In this study we let $L = 172$. This way, an LV is fully captured by the x-y-z coordinates of each landmark. Written in vector form, the LV landmark set or pose (we don't distinguish these two terms hereinafter) is $\boldsymbol{p} = (x_1, y_1, z_1, \ldots, x_L, y_L, z_L)^\top \in \mathbb{R}^{3L}$. Given a 3D CT volume $I \in \mathbb{R}^{a \times b \times c}$ with the x-y-z size $a, b, c$ respectively, we directly predict the LV by a trained regressor $R(\cdot) : \mathbb{R}^{a \times b \times c} \mapsto \mathbb{R}^{3L}$:

$$\boldsymbol{p} = R(I). \qquad (1)$$

Considering the various poses that an LV can have, learning the regressor $R(\cdot)$ is not easy. Another difficulty is that most successful features in computer vision that a regressor can rely on are unfortunately pose-blind, i.e., they just depend on the image pixel values. To address the issues, Dollar et al [3] proposes that the pose is fitted in a *multi-stage*, gradually refined way, where the feature pool at each stage explicitly depend on the current pose estimation up to that stage. This methodology is called Cascaded Pose Estimation[3] and has been successfully applied to applications such as face alignment[1]. Specifically, the regressor (1) is written as an ensemble of $T \in \mathbb{N}^+$ regressors:

$$R(I) = \sum_{t=1}^{T} R^t(I; \boldsymbol{p}^{t-1}), \qquad (2)$$

where the pose is updated recursively:

$$\boldsymbol{p}^t = \boldsymbol{p}^{t-1} + R^t(I; \boldsymbol{p}^{t-1}) \qquad (3)$$

for *stage* $t = 1, 2, ..., T$. The $\boldsymbol{p}^0$ is just an initial guess. Note that $R(I; \boldsymbol{p}^{t-1})$, referred to as *stage regressor*, depends on both the image and the pose estimation up to $\boldsymbol{p}^{t-1}$. This dependency is done via the Pose Index Feature. In the following sub sections we give greater details.

**Pose Indexed Feature and Random Pixel Difference.** Many effective features in computer vision, such as Haar wavelet, SIFT descriptor, Histogram of Gradients (HOG), are computed in a way which involves only the pixel values in the image, no matter what the true pose is for the interested object in the image. This makes those features vulnerable to pose variation. To build reliable features for pose estimation, Fleuret et al [5] proposes the concept of Pose Indexed Feature, which means the feature extraction explicitly depends on the pose itself and should hopefully be invariant to changes in pose. The most straightforward implementation of such a feature is to extract the conventional feature (e.g., Haar, SIFT, HOG...) in just a small neighborhood of every landmark and then to concatenate them, as suggested in Dollar et al [3] and Cao et al [1]. This choice has also been mathematically proven to be pose invariant under mild assumption [3].

In this paper, we follow the line of Dollar et al [3] Cao et al [1] and have the Pose Index Features built on top of the Random Pixel Difference [8]. To be precise, for the $i$-th landmark of our LV shape, we let it be the origin of a local coordinate system, in which we uniform randomly generate $M$ point pairs within a circle with radius $r \in \mathbb{R}$. Let $v_{ij}$ be the pixel value difference for the $j$-th pair, we build the features $\boldsymbol{x}$ by concatenating all $v_{ij}$:

$$\boldsymbol{x}^t = (v_{ij})^\top, \qquad (4)$$

for $i = 1, ..., L, j = 1, ..., L$ so that $\boldsymbol{x}^t \in \mathbb{R}^{ML}$, where the superscript $t$ emphasize that the "life cycle" of the feature pool is only the $t$-th stage.

**Learning the Stage Regressor** The stage regressor is added sequentially in a forward greedy way. At stage $t$, the target to be regressed is the pose residual $\Delta \boldsymbol{p} = \boldsymbol{p} - \boldsymbol{p}^t$ between the ground truth pose $\boldsymbol{p}$ and the pose estimation at stage $t$. Following the work of previous studies [7], we adopt a Least Square Boosting Tree ensemble trained over the Random Pixel Difference features $\boldsymbol{x}^t$:

$$R_t(\boldsymbol{x}) = \sum_{k=1}^{K} \nu f_t^k(\boldsymbol{x}^t), \qquad (5)$$

where $K \in \mathbb{N}^+$ is the number of trees, $f_t^k(\cdot) : \mathbb{R}^D \mapsto \mathbb{R}^{3L}$ is a vector-valued regression tree that each leaf outputs a vector with $3L$ components, and $\nu \in (0, 1)$ is the step size (a.k.a. shrinkage factor) controlling the learning rate [6].

Unlike the PCA used in ASM, the regressor (5) enforces the shape constraint by sharing the features and the vector-valued prediction, as is analyzed in Cao et al [1].

**Training Data Augmentation and Initial Guess** Having described the fashion of how to learn the stage regressor, we can write the training data with size $N$ as a set of triplets: $\{(I_i, \boldsymbol{p}_i^0, \Delta\boldsymbol{p}_i)\}_{i=1}^{N}$, where $I_i$ is the image, $\boldsymbol{p}_i^0$ is the initial guess and the $\Delta\boldsymbol{p}_i$ is the residual between the guess and the ground truth. The $\Delta\boldsymbol{p}_i$ is reduced and updated at each stage with the addition of new stage regressor.

To improve the quality of regressor (2) even when $N$ is small (e.g., several hundreds), previous studies [3] [1] propose a data augmentation trick. This is done by sampling other training instance's ground truth as the initial guess. This way, a new training data set $\{I_j, \boldsymbol{p}_j^0, \Delta\boldsymbol{p}_j\}_{j=1}^{N^*}$ can be obtained with $N^* \gg N$. Note that the number of unique images remains, but many $\{\boldsymbol{p}_j^0, \Delta\boldsymbol{p}_j\}$ pairs are made.

In testing stage for any unseen volume $I$, we feed $I$ into equation (2) and simply let the initial guess $\boldsymbol{p}^0$ be the mean shape of those $\boldsymbol{p}_j^0$, $j = 1, ..., N^*$ over the augmented training data.

## 3 Experiments

**Experiment Setup.** We verify the proposed algorithm on a dataset including 139 cardiac CT volumes from 139 individual patients. The in-slice size (x-size and y-size) is always $512 \times 512$, but the number of slices (z-size) ranges from 153 to 357. For all the CT volumes, the in-slice resolution is isotropic (i.e., always the same for x- and y- axis), ranging from 0.28 to 0.49 millimeters; while the slice thickness (i.e., resolution for z-axis) ranges from 0.30 to 0.63 millimeters.

**Accuracy.** We follow the Zheng et al method[10] for the accuracy measure of the algorithm, i.e., adopting *symmetric surface-surface distance* (called point-to-mesh distance in Zheng et al[10]). We report the symmetric surface-surface distance $d(\hat{\boldsymbol{p}}, \boldsymbol{p})$ between the predicted pose $\hat{\boldsymbol{p}}$ and the ground truth pose $\boldsymbol{p}$ via 5-fold cross validation. In Table 1 we give the mean and the standard deviation (in parenthesis) for both the inner surface and the outer surface of LV. As a comparison, we also cite the results from Zheng et al [10] experiments on their own dataset of 457 CT volumes. See Zheng et al [10] for detailed setup. In regards to accuracy, our algorithm is on par with Zheng et al [10]; But our algorithm is more stable since it has lower deviation than Zheng et al [10].

**Testing Speed.** We implement the testing stage of the regressor (2) in C++ (the video demo for our software is available at `https://youtu.be/t0gbc026Hl4`). To measure how fast the testing is, we run the algorithm 500 times on several different CT volumes and report the average time as in Table 1. We also cite the Zheng et al [10] result where only the total time for all of the four chamber cases is reported and the separate timing for LV is unavailable. Even when multiplied four times, the speedup of our algorithm over [10] is still in the order of Zheng et al $\times 1000$.

Table 1. The segmentation error (millimeters) and the running time (second).

| This paper | | | Zheng et al[10] | | |
|---|---|---|---|---|---|
| Outer-Err | Inner-Err | Running-Time* | Outer-Err | Inner-Err | Running-Time** |
| 1.23(0.11) | 1.21(0.11) | $1.2 \times 10^{-3}$ | 1.21(0.41) | 1.13(0.55) | 4.0 |

* Only for LV (3.4G Hz CPU);   ** For all the four chambers (3.2G Hz CPU)

### 3.1 Parameters

The main tuning parameters of the proposed algorithm are listed below. $N^*$: the augmented training data size; $T$: number of stages; $K$: number of trees in each stage; $d$: the tree depth; $M$: number of features extracted in a point's neighborhood; $\nu$: the step size (called shrinkage factor in Boosting regression literature [7]) when building regression tree ensemble in each stage. Our result in Table 1 is obtained with $N^* = 62,500$, $T = 50$, $K = 10$, $d = 3$, $M = 5$, $\nu = 0.2$.

To investigate how the tuning parameters affect the final result, we set a smaller dataset, called *dataset II* hereafter, due to the constraint of computational resource. Specifically, we randomly pick 95 cases for training and 29 cases for testing to constitute dataset II.

We find that tuning $d$ and $M$ around $d = 3$ and $M = 5$ doesn't change the result on dataset II significantly.

The data augmentation trick proposed in previous studies [3][1] improves the result in our problem. In Table 2 we report the results on dataset II with varying $N^*$. However, the result becomes flat over $N^* = 62500$ in the experiment.

Table 2. Segmentation error (in millimeters) with varying $N$ or $N^*$ on dataset II.

| | $N^*$ with $N = 139$ | | | | | | $N$ with $N^* = 2500$ | | | |
|---|---|---|---|---|---|---|---|---|---|---|
| | 100 | 500 | 2,500 | 12,500 | 62,500 | 312,500 | 10 | 30 | 60 | 95 |
| Outer-Err | 3.99 | 1.75 | 1.28 | 1.16 | 1.22 | 1.22 | 3.61 | 1.61 | 1.38 | 1.28 |
| Inner-Err | 3.90 | 1.73 | 1.26 | 1.15 | 1.15 | 1.17 | 3.55 | 1.58 | 1.33 | 1.26 |

We also confirm the effectiveness of cascaded pose regression as is reported in previous studies [3][1][7]. For example, we fix the total number of trees $T'$ ($= T \times K$) to be 500 and let $T = 1$ and $K = 500$, in which case the cascaded pose regression reduces to regular regression since there is only one stage. When $N^* = 2500$ and other parameters unchange, the result on dataset II degrades to 4.98 for outer surface and 4.95 for inner surface (both in millimeters).

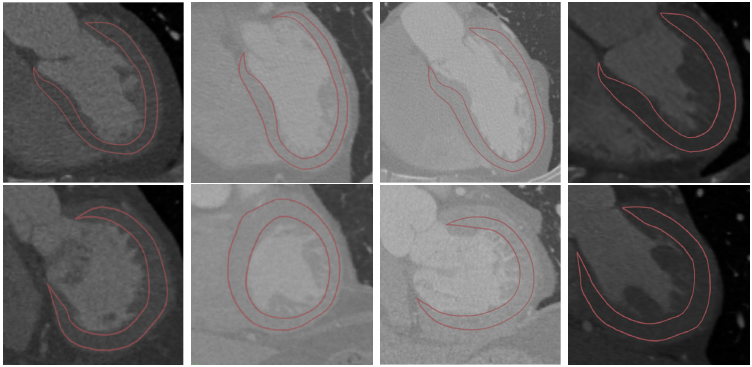

**Fig. 3.** Examples of the proposed segmentation algorithm.

### 3.2 Effect of Training Data Size

For the general computer vision problems involving tasks like face pose estimation, it is common to have tens of thousands of annotated training data, e.g., in Cao et al [1] the training data size is approximately 10K.

However, CT volume data are difficult to collect, not mentioning that the annotation is time consuming since a human expert has to work on 3D volume. Typical training data size in medical image literature is about several hundreds [10][12][4]. In this paper the number of training CT volumes is of the same order. Although the training data size is far less than that in general computer vision problems, still, acceptable results seemed to be obtained with such a "small dataset". It's thus interesting to investigate *1) How will the training data size affect the final result? In particular, is it possible to train on only 10 CT volumes and obtain a comparable result when training on 100 CT volumes?*

As described in Section 2, in order to improve the performance of Cascaded Pose Regression, a training data augmentation trick was introduced [3][1] where the training data size is scaled up by $\times$ 20, which is also adopted in our implementation. It is natural to ask *2) Will the data augmentation trick alone relieve the suffering of insufficient training data in our problem?*

To address both of the questions, we randomly sub-sample the original training data so that the size of the sub training set is $N = 10, 30, 60$. In Table 2, we show the results on dataset II for the regressor (2) trained with different $N$ but tested on the same testing set, for which we fix $N^* = 2500$. It is clear that the result of the proposed algorithm improves gradually with increasing $N$. On the other hand, the results improve with larger $N^*$ with fixed $N = 139$, but they get flat beyond certain $N^*$. To conclude, even if the annotation in Cardiac CT is time consuming, we should still make the effort to obtain more annotated data, since the data augmentation trick alone cannot relive the suffering of insufficient training data.

## 4  Conclusion

In this paper we propose a fast method to segment the left ventricles from images. The average time taken per case is 1.2 millisecond, which is almost three orders of magnitude improvement over several seconds reported in the previous studies. The proposed method is more stable (lower standard deviation by 5-fold cross validation) and doesn't significantly compromise the accuracy. We present a couple of examples of the proposed algorithm in Figure 3. Due to the generalizability of our approach, we will investigate the feasibility and performance when applied to the segmentation of all four heart chambers and other organ structures (e.g., liver, kidney) from images by different modalities (e.g. MR and ultrasound).

## References

1. Cao, X., Wei, Y., Wen, F., Sun, J.: Face alignment by explicit shape regression. International Journal of Computer Vision 107(2), 177–190 (2014)
2. Criminisi, A., Shotton, J., Robertson, D., Konukoglu, E.: Regression forests for efficient anatomy detection and localization in ct studies. In: Medical Computer Vision. Recognition Techniques and Applications in Medical Imaging, pp. 106–117. Springer (2011)
3. Dollár, P., Welinder, P., Perona, P.: Cascaded pose regression. In: Computer Vision and Pattern Recognition (CVPR), 2010 IEEE Conference on. pp. 1078–1085. IEEE (2010)
4. Ecabert, O., Peters, J., Schramm, H., Lorenz, C., von Berg, J., Walker, M.J., Vembar, M., Olszewski, M.E., Subramanyan, K., Lavi, G., et al.: Automatic model-based segmentation of the heart in ct images. Medical Imaging, IEEE Transactions on 27(9), 1189–1201 (2008)
5. Fleuret, F., Geman, D.: Stationary features and cat detection. Journal of Machine Learning Research 9(2549-2578), 16 (2008)
6. Friedman, J., Hastie, T., Tibshirani, R., et al.: Additive logistic regression: a statistical view of boosting (with discussion and a rejoinder by the authors). The annals of statistics 28(2), 337–407 (2000)
7. Kazemi, V., Josephine, S.: One millisecond face alignment with an ensemble of regression trees (2014)
8. Ozuysal, M., Calonder, M., Lepetit, V., Fua, P.: Fast keypoint recognition using random ferns. Pattern Analysis and Machine Intelligence, IEEE Transactions on 32(3), 448–461 (2010)
9. Viola, P., Jones, M.J.: Robust real-time face detection. International journal of computer vision 57(2), 137–154 (2004)
10. Zheng, Y., Barbu, A., Georgescu, B., Scheuering, M., Comaniciu, D.: Four-chamber heart modeling and automatic segmentation for 3-d cardiac ct volumes using marginal space learning and steerable features. Medical Imaging, IEEE Transactions on 27(11), 1668–1681 (2008)
11. Zhou, S.K.: Shape regression machine and efficient segmentation of left ventricle endocardium from 2d b-mode echocardiogram. Medical image analysis 14(4), 563–581 (2010)
12. Zhou, S.K., Comaniciu, D.: Shape regression machine. In: Information Processing in Medical Imaging. pp. 13–25. Springer (2007)